# Optimizing Hyper parameters in CNN for Soil Classification using PSO and Whale Optimization Algorithm

Yasir Noor
yasir.altai@uomosul.edu.iq

Fawziya Mahmood Ramo
fawziyaramo@uomosul.edu.iq

College of Computer Science and Mathematics Center

Remote Sensing

Mosul University Mosul, Iraq

Mosul University Mosul, Iraq

Mahmood Siddeeq Qadir

Muna Jaffer Al-Shamdeen

Mahmoodsq@uomosul.edu.iq
muna.jaffer@uomosul.edu.iq

Remote Sensing Cente

College of Computer Science and Mathematics

Mosul University Mosul, Iraq

Mosul University Mosul, Iraq

## Abstract

Classifying soil images contributes to better land management, increased agricultural output, and practical solutions for environmental issues. The development of various disciplines, particularly agriculture, civil engineering, and natural resource management, is aided by understanding of soil quality since it helps with risk reduction, performance improvement, and sound decision-making . Artificial intelligence has recently been used in a number of different fields. In this study, an intelligent model was constructed using Convolutional Neural Networks to classify soil kinds, and machine learning algorithms were used to enhance the performance of soil classification .

To achieve better implementation and performance of the Convolutional Neural Networks algorithm and obtain valuable results for the process of classifying soil type images, swarm algorithms were employed to obtain the best performance by choosing Hyper parameters for the Convolutional Neural Networks network using the Whale optimization algorithm and the Particle swarm optimization algorithm, and comparing the results of using the two algorithms in the process of multiple classification of soil types. The Accuracy and F1 measures were adopted to test the system, and the results of the proposed work were efficient result .

1- Introduction

Soil classification is a critical task in agriculture, geotechnical engineering, and environmental management. Accurate classification helps in determining soil properties, fertility levels, and suitability for various applications, such as crop selection, construction, and land-use planning [1]. Traditional soil classification methods rely on laboratory analyses, which can be time-consuming and costly. With advancements in artificial intelligence (AI), machine learning (ML), and deep learning (DL), automated soil classification has gained significant attention [2].

Among deep learning models, Convolutional Neural Networks (CNNs) have shown remarkable success in image-based classification tasks, making them highly effective for soil classification based on spectral, texture, and microscopic soil images [3]. However, CNN performance depends heavily on hyperparameter selection, such as learning rate, filter size, and the number of layers. Manual tuning or grid search methods often lead to suboptimal results and require extensive computational resources [4].

To address this challenge, metaheuristic optimization algorithms such as the Whale Optimization Algorithm (WOA) and Particle Swarm Optimization (PSO) have been employed to fine-tune CNN hyperparameters efficiently [5][6]. WOA, inspired by the social hunting behavior of humpback whales, and PSO, based on the collective intelligence of bird flocks and fish schools, provide powerful global search capabilities to optimize CNN architectures. By integrating these optimization techniques, CNN models can achieve improved accuracy, robustness, and generalization for soil classification tasks [7].

This research explores the effectiveness of WOA and PSO in optimizing CNN hyperparameters for soil classification. The study evaluates and compares the performance of optimized CNN models using various soil datasets, highlighting the advantages of metaheuristic-based hyperparameter tuning in deep learning applications. The findings contribute to enhancing automated soil classification, aiding precision agriculture, environmental monitoring, and geotechnical assessments [8].

## 2-Literature Review

Chetan Raju et al., 2024[9]: In this study, transfer learning was applied to a dataset of soil images using a hybrid approach to the ResNet50 algorithm and improving its performance by modifying the layers of the transfer learning algorithm. The total number of images used for the dataset was 720 images. This data was collected from several public sources. The research methodology was to collect soil images of four main types (Alluvial, Red, Black, Clay), where the results of classification accuracy were Accuracy: 94.3%.

Satheesh Mokkapati and et al., 2024 [10]: In this research paper, soil images are classified using deep learning techniques with high accuracy by developing a deep learning model using convolutional neural networks (CNNs), The research methodology was to collect soil images from multiple sources and use data augmentation techniques such as rotation and zooming and apply a deep learning model, The following results were obtained: Accuracy: 93.1%.

S.Yaswanth Kiran and et al., 2024[11]: The research paper presented a system for automatic soil classification using a system based on convolutional neural networks (CNN) using a dataset of digital images instead of the traditional manual effort and longtime used to classify soil. The system provides accuracy and speed in soil classification, especially supporting decision makers for use in land management and the field of agriculture, Samples of soil images were collected, and then image processing operations such as normalization and data enhancement were performed, The classification accuracy was as follows - Accuracy: 92.5%.

P Gopala Raju and et al., 024[12]: The study, among its objectives, reviews recent developments in soil classification methodologies, as studies related to soil classification through digital images have focused on the integration of innovative machine learning algorithms, Data processing techniques strategies such as missing data compensation and dealing with class imbalance have been discussed. The most important results of the algorithms were shown as follows: K-Nearest Neighbors (KNN) algorithm: classification accuracy of up to 95%. Support Vector Machines (SVM) algorithm :classification accuracy of up to 94%. The decision Tree algorithm: has a classification accuracy of up to 90% And the Random Forest algorithm: Shows the highest classification accuracy of up to 97%.

**3-Optimizing Hyper parameters**

Hyperparameters play a crucial role in the performance and efficiency of machine learning models, especially deep learning architectures like Convolutional Neural Networks (CNNs). Unlike model parameters, which are learned from data during training, hyperparameters are predefined settings that control the learning process and influence the model's ability to generalize to new data. Key hyperparameters in CNNs include the learning rate, batch size, number of layers, filter sizes, activation functions, and dropout rates. Selecting the optimal combination of hyperparameters is essential to avoid underfitting or overfitting, ensuring the model achieves high accuracy and robustness[13][14]

Traditional hyperparameter tuning methods, such as grid search and random search, often require extensive computational resources and time. To overcome these limitations, metaheuristic optimization algorithms like Particle Swarm Optimization (PSO) and Whale Optimization Algorithm (WOA) have gained popularity. These algorithms efficiently explore the hyperparameter space by balancing exploration and exploitation, leading to faster convergence and improved model performance [15].

6- Paper Methodology

In this research, an intelligent model was built to classify soil types, including four types of soil (Alluvial, Black, Clay and Red) using a CNN network, with its reconstruction and configuration, and focusing on choosing the optimal hyper parameters, this was done using optimization algorithms such as the whale algorithm and PSO, as shown in the following figure (1).

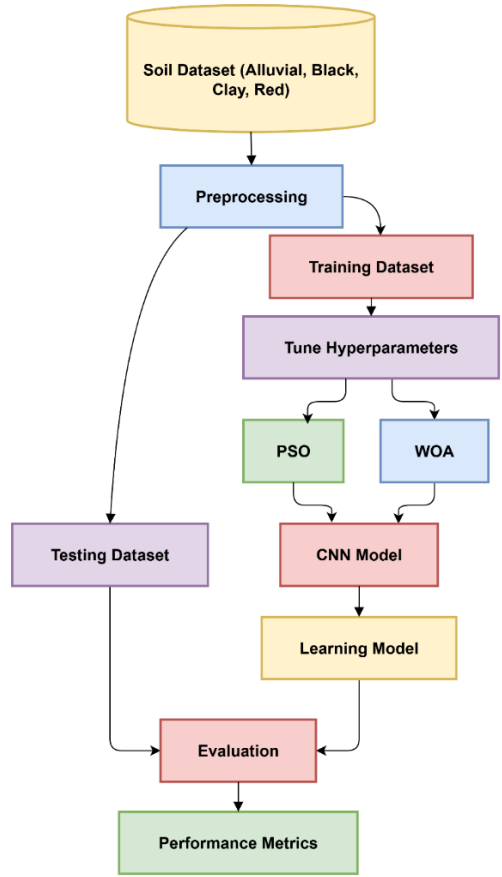

Figure 1: Block diagram for system

6-1 Soil Dataset Preparation and Preprocessing

Before using it, the dataset is prepared and preprocessed to ensure compatibility with the CNN model.   In this research, a soil data set was adopted form Kaggle (www.kaggle.com/datasets/naivedatamodel/soil-dataset) , total number of images was 340 images ,The following steps are performed:

Loading the Dataset: The dataset divided into training and testing datasets. Four categories of soil images are included. Images were then resized to a uniform dimension (300 *300 pixels) and organized into 32 batches. The following figure represent sample of soil images

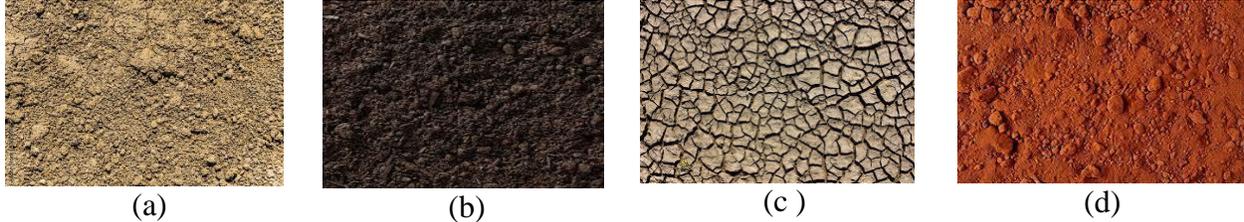

(a)     (b)     (c )     (d)

Figure-2 sample of  soil images :a-Alluvial  b-Black   c-Clay     d-Red

Label Encoding: Labels are encoded in categorical format to facilitate multi-class classification.
Normalization: Pixel values are normalized to the range [0, 1] using a Rescaling layer to improve model convergence and stability

6-2 Search Space:

Four hyperparameters are applied to optimize the proposed CNN model, these hyperparameters are defined as follows:

| Hyper-parameter | Symbol | Type | Range Considered |
|---|---|---|---|
| convolutional filters | $N_f$ | integer | [8, 32] |
| dense units | $N_d$ | integer | [32, 128] |
| Dropout rate | $p$ | continuous | [0.1, 0.5] |
| Learning rate | $\eta$ | continuous | $1\times10^{-4}, 1\times10^{-2}$ |

6-3 CNN Architecture Design

To remain amenable to hyper-parameter search, we adopt a **lean yet expressive** CNN backbone whose sole architectural degrees of freedom are (i) the **number of convolutional filters** $N_f$, (ii) the **dense-layer width** $N_d$, and (iii) the **dropout probability** $p$. The learning-rate $\eta$ of the Adam optimizer constitutes the fourth tunable scalar. The network processes 300×300×3 RGB soil images through the following feed-forward stack:

- Conv2D(32 filters, 3×3, ReLU)
- MaxPooling2D(2×2)
- Flatten
- Dense($N_d$, ReLU)
- Dropout($p$)
- Dense(4, Softmax)

The complete schematic is illustrated in **Figure 3**. All convolutional kernels are initialized with He-normal weights, and batch-normalization is omitted to keep the hyper-parameter count minimal.

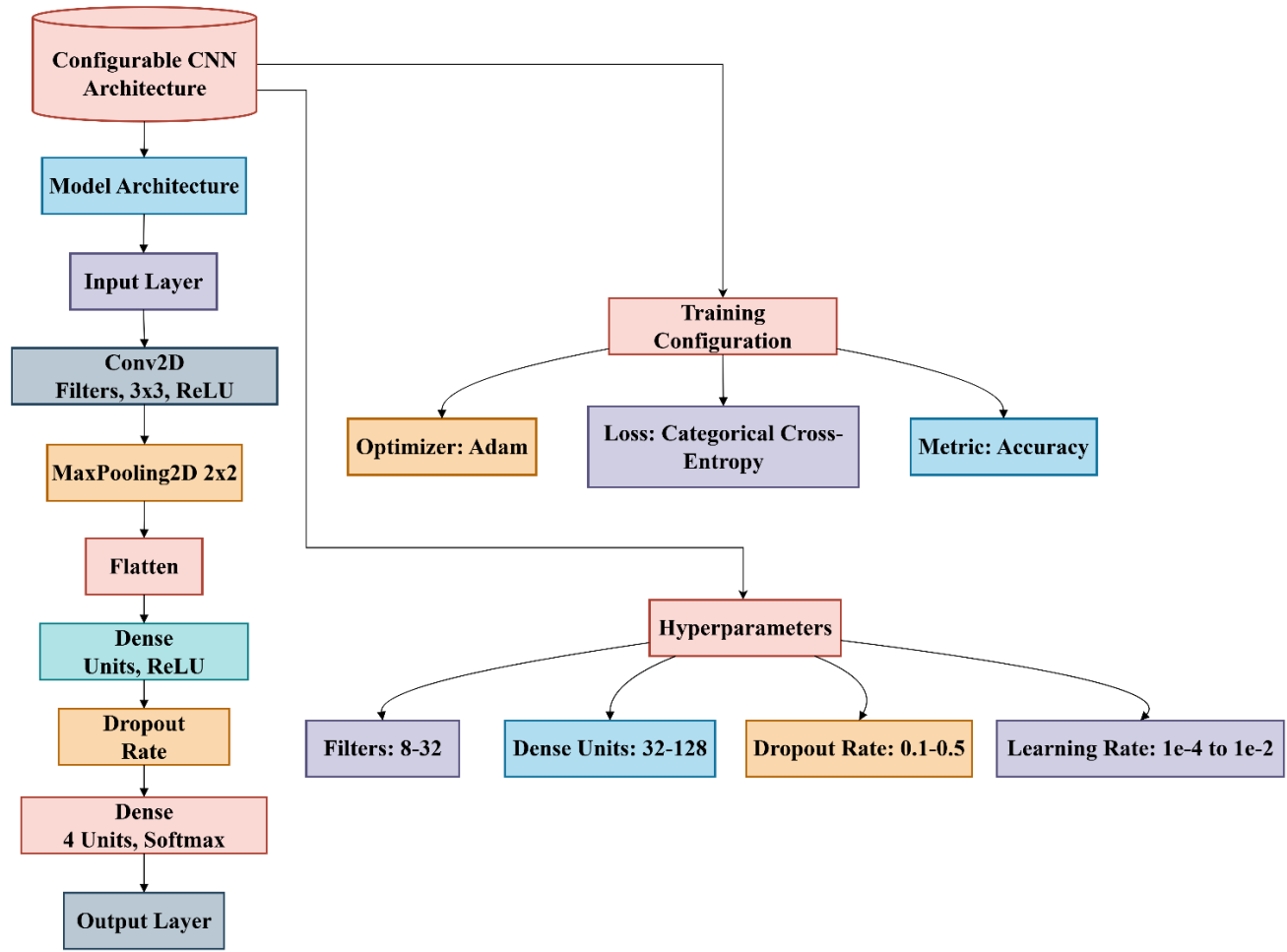

## 6-5 PSO-CNN Models

Particle Swarm Optimization (PSO) optimizes CNN hyperparameters by simulating social behavior of bird flocks. Five particles represent different hyperparameter configurations, each maintaining position (current parameters) and velocity (change direction). Through 5 iterations, particles adjust their velocities based on personal best (pBest) and global best (gBest) experiences, balancing exploration and exploitation. Each particle undergoes rapid 5-epoch CNN training for fitness evaluation using error = 1 − accuracy, with continuous updates to personal and global best solutions.

ALGORITHM PSO-CNN_Optimization

```
BEGIN
  // Phase 1: Swarm Initialization
  FOR particle_id = 1 TO swarm_size DO
     position[particle_id] = random_sample(search_space)
     velocity[particle_id] = random_initialize_velocity()
     model = train_CNN(position[particle_id], epochs=5)
     fitness[particle_id] = 1 - evaluate_model(model, test_data)
     pbest[particle_id] = position[particle_id]
```

```
      pbest_fitness[particle_id] = fitness[particle_id]
   END FOR

   gbest = argmin(pbest_fitness)
   gbest_fitness = min(pbest_fitness)

   // Phase 2: Iterative Optimization
   FOR iteration = 1 TO max_iterations DO
      FOR each particle_id DO
         // Generate random coefficients
         r1 = random_uniform(0, 1)
         r2 = random_uniform(0, 1)

         // Velocity update equation
         velocity[particle_id] = w * velocity[particle_id] +
                     c1 * r1 * (pbest[particle_id] - position[particle_id]) +
                     c2 * r2 * (gbest - position[particle_id])

         // Position update
         position[particle_id] = position[particle_id] + velocity[particle_id]

         // Boundary constraint enforcement
         position[particle_id] = clip_to_bounds(position[particle_id], search_space)

         // Model evaluation
         model = train_CNN(position[particle_id], epochs=5)
         current_fitness = 1 - evaluate_model(model, test_data)

         // Personal best update
         IF current_fitness < pbest_fitness[particle_id] THEN
            pbest[particle_id] = position[particle_id]
            pbest_fitness[particle_id] = current_fitness
         END IF

         // Global best update
         IF current_fitness < gbest_fitness THEN
            gbest = position[particle_id]
            gbest_fitness = current_fitness
         END IF
      END FOR
   END FOR

   RETURN gbest, final_model
END
```

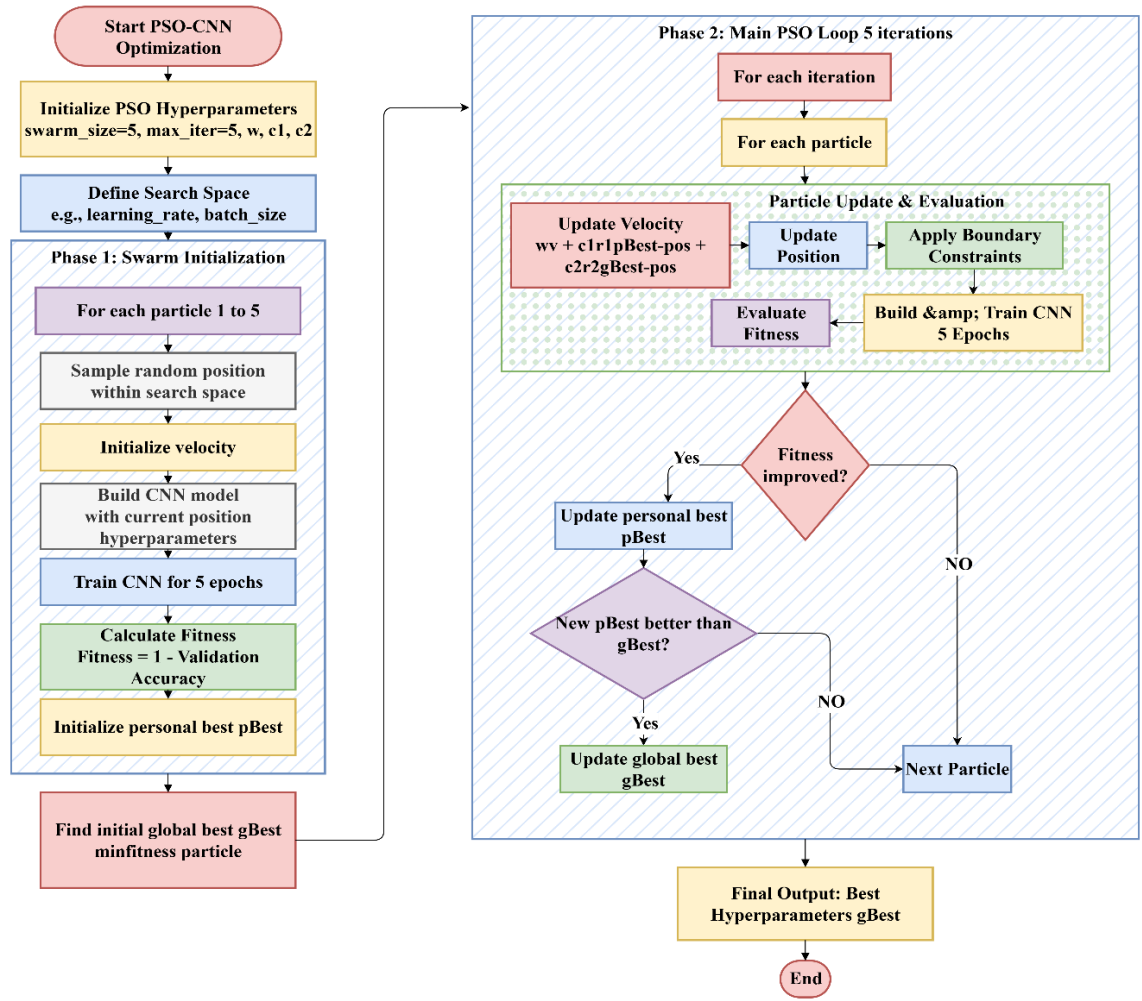

## 6-4 WOA-CNN Models

The Whale Optimization Algorithm (WOA) optimizes CNN hyperparameters by simulating whale hunting behavior. The process begins by creating a population of five whale solutions, each representing different CNN hyperparameter configurations randomly sampled from a predefined search space. Each whale undergoes rapid evaluation through 5-epoch training sessions, with fitness calculated as error = 1 − accuracy. Through 10 optimization iterations, whales update their positions using three behavioral patterns: shrinking encircling (moving toward the current best solution when |A| < 1), spiral bubble-net feeding (following a logarithmic spiral path), and exploration (searching randomly when |A| ≥ 1). The algorithm continuously tracks and updates the best-performing whale while ensuring all hyperparameters remain within valid bounds.

WOA-CNN Models - Pseudo Code

```
BEGIN
    FOR whale_id = 1 TO population_size DO
        hyperparameters[whale_id] = random_sample(search_space)
        model = train_CNN(hyperparameters[whale_id], epochs=5)
        accuracy = evaluate_model(model, test_data)
        fitness[whale_id] = 1 - accuracy
    END FOR
```

```
      best_solution = argmin(fitness)

   FOR iteration = 1 TO max_iterations DO
      FOR each whale_id DO
         A = calculate_coefficient_A(iteration)
         C = random_uniform(0, 2)

         IF |A| < 1 THEN
            new_position = best_solution - A * |C * best_solution - current_position|
         ELSE
            random_whale = select_random_whale()
            new_position = random_whale - A * |C * random_whale - current_position|
         END IF

         spiral_factor = exp(b * l) * cos(2π * l)
         new_position = |best_solution - current_position| * spiral_factor + best_solution

         new_position = clip_to_bounds(new_position, search_space)

         model = train_CNN(new_position, epochs=5)
         new_fitness = 1 - evaluate_model(model, test_data)

         IF new_fitness < best_fitness THEN
            best_solution = new_position
            best_fitness = new_fitness
         END IF
      END FOR
   END FOR

   RETURN best_solution, final_model
END
```

## 6-4 WOA-CNN Models

The Whale Optimization Algorithm (WOA) optimizes CNN hyperparameters by simulating whale hunting behavior. The process begins by creating a population of five whale solutions, each representing different CNN hyperparameter configurations randomly sampled from a predefined search space. Each whale undergoes rapid evaluation through 5-epoch training sessions, with fitness calculated as error = 1 − accuracy. Through 10 optimization iterations, whales update their positions using three behavioral patterns: shrinking encircling (moving toward the current best solution when $|A| < 1$), spiral bubble-net feeding (following a logarithmic spiral path), and exploration (searching randomly when $|A| \geq 1$). The algorithm continuously tracks and updates the best-performing whale while ensuring all hyperparameters remain within valid bounds.

WOA-CNN Models - Pseudo Code

```
BEGIN
   FOR whale_id = 1 TO population_size DO
      hyperparameters[whale_id] = random_sample(search_space)
      model = train_CNN(hyperparameters[whale_id], epochs=5)
      accuracy = evaluate_model(model, test_data)
```

```
      fitness[whale_id] = 1 - accuracy
   END FOR

   best_solution = argmin(fitness)

   FOR iteration = 1 TO max_iterations DO
      FOR each whale_id DO
         A = calculate_coefficient_A(iteration)
         C = random_uniform(0, 2)

         IF |A| < 1 THEN
            new_position = best_solution - A * |C * best_solution - current_position|
         ELSE
            random_whale = select_random_whale()
            new_position = random_whale - A * |C * random_whale - current_position|
         END IF

         spiral_factor = exp(b * l) * cos(2π * l)
         new_position = |best_solution - current_position| * spiral_factor + best_solution

         new_position = clip_to_bounds(new_position, search_space)

         model = train_CNN(new_position, epochs=5)
         new_fitness = 1 - evaluate_model(model, test_data)

         IF new_fitness < best_fitness THEN
            best_solution = new_position
            best_fitness = new_fitness
         END IF
      END FOR
   END FOR

   RETURN best_solution, final_model
END
```

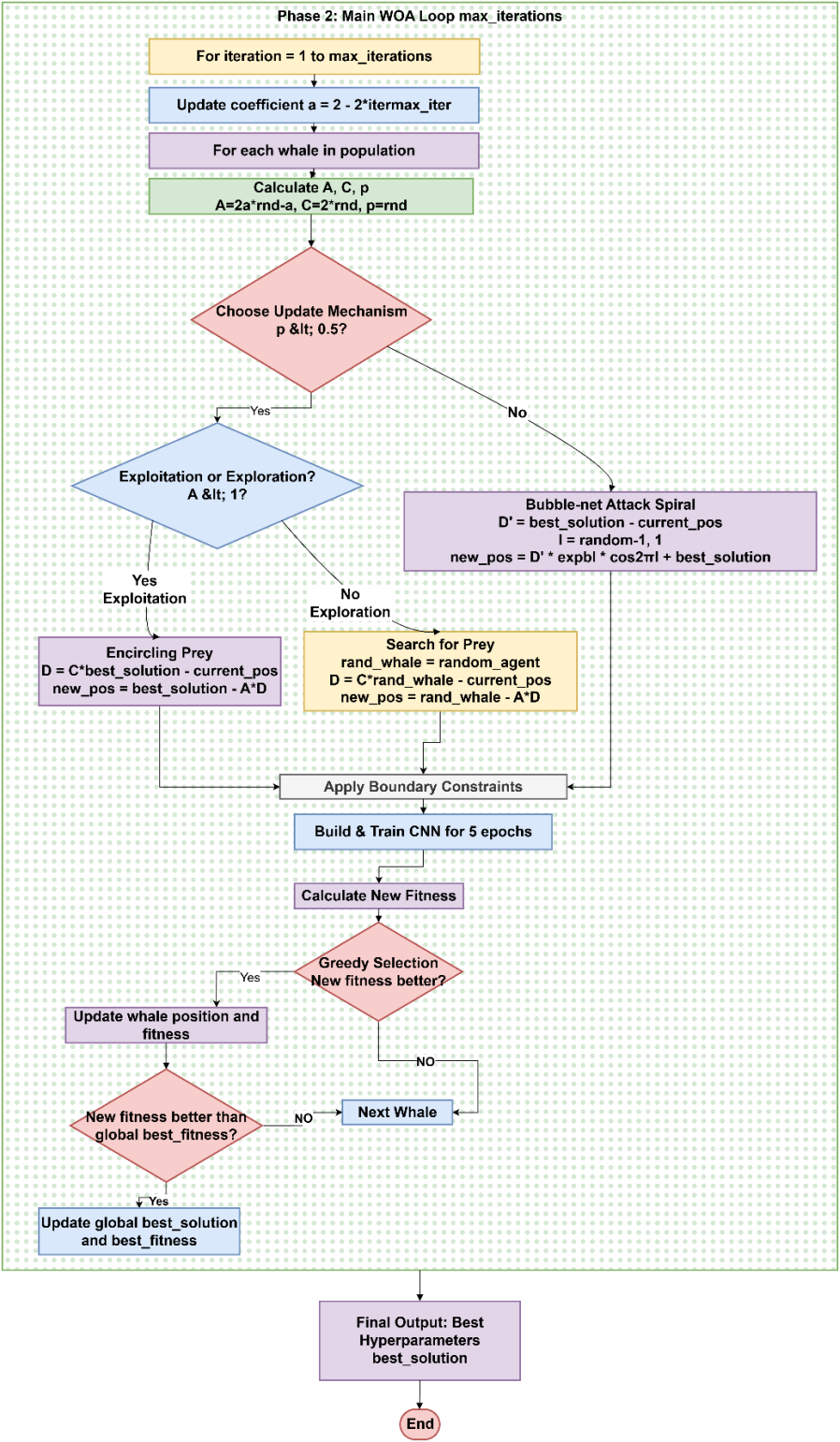

7- Results and Discussion

The range of hyperparameters (Num_filters , Dense_unit, Dropout_rate, learning rate) are initialized to carry out the training processes on the dataset as shown in table (1). The most important hyperparameters are found using WOA and PSO optimization techniques, table 2 shows the most important hyperparameters after implementing both algorithms.

Table (1): The Hyperparameter Range for both the WOA-CNN and PSO-CNN Models

| Hyperparameters | Range |
|---|---|
| Num_filters | [8 – 32] |
| Dense_unit | [32- 128] |
| Dropout_rate | [0.1-0.5] |
| Learning rate | [0.0001-0.01] |

Table (2): Best Hyperparameters Values of WOA-CNN and PSO-CNN Models

| Hyperparameters | WOA-CNN | PSO-CNN |
|---|---|---|
| Num_filters | 8 | 9 |
| Dense_unit | 51 | 88 |
| Dropout_rate | 0.3321050934710237 | 0.13298126723690565 |
| Learning rate | 0.009392635740413055 | 0.005372761052242158 |

The WOA-CNN and PSO-CNN models were trained and evaluated using the Soil Dataset. The performance results of WOA-CNN and PSO-CNN in Soil Dataset is illustrated in table 3 which shows the metrics value of precision, recall and F1-score for each class and shows the metrics value of overall accuracy terms of the evaluation metrics (precision , recall, f1-score and accuracy) for WOA-CNN and PSO-CNN in Soil Dataset as shown in table (3), It is observed that both models achieved high results and the PSO-CNN model has produced the highest value of accuracy comparing with WOA-CNN due to the difference in the structure of both algorithm (WOA and PSO), which means the pos has best performance in optimizing hyperparameter.

Table (3): Evaluation Results of WOA-CNN and PSO-CNN on Soil Dataset

| Evaluation Metrics of WOA-CNN on Soil Dataset | | | |
|---|---|---|---|
| Classes | Precision | Recall | F1-score |
| Alluvial soil | 0.94 | 0.83 | 0.88 |
| | Black Soil | 0.92 | 0.98 | 0.95 |
| In | Clay soil | 0.89 | 0.86 | 0.88 |
| | Red soil | 0.98 | 0.98 | 0.98 |
| Accuracy of WOA-CNN on Soil Dataset: 94 ||||
| Evaluation Metrics of PSO-CNN on Soil Dataset ||||
| Classes | Precision | Recall | F1-score |
| Alluvial soil | 1.00 | 0.85 | 0.92 |
| Black Soil | 0.97 | 0.99 | 0.98 |
| Clay soil | 0.87 | 1.00 | 0.93 |
| Red soil | 1.00 | 0.97 | 0.99 |
| Accuracy of PSO -CNN on Soil Dataset: 96 ||||

8- Conclusions

This study demonstrated the effectiveness of using metaheuristic optimization techniques—namely the Whale Optimization Algorithm (WOA) and Particle Swarm Optimization (PSO)—for fine-tuning the hyperparameters of Convolutional Neural Networks (CNNs) in the task of soil image classification. By systematically optimizing critical hyperparameters such as the number of filters, dense units, dropout rate, and learning rate, both approaches achieved high accuracy and robustness on a multi-class soil dataset consisting of Alluvial, Black, Clay, and Red soil types.

The experimental results revealed that both WOA-CNN and PSO-CNN models performed strongly, with accuracies of 94% and 96%, respectively. The PSO-CNN model consistently outperformed WOA-CNN across evaluation metrics including precision, recall, and F1-score, particularly in the classification of Black and Red soil types. These findings indicate that PSO is more efficient in exploring and exploiting the hyperparameter search space, leading to superior optimization performance compared to WOA.

The confusion matrix analysis further confirmed the reliability of both models, showing high prediction accuracy across all soil classes with minimal misclassifications. Nevertheless, certain confusions between visually similar soil categories highlight the need for larger and more diverse datasets to improve feature separability.

In summary, the integration of swarm intelligence algorithms with CNNs provides a powerful framework for automated soil classification, reducing reliance on manual feature engineering and costly laboratory analyses. The proposed models not only enhance classification accuracy but also contribute to precision agriculture, sustainable land management, and environmental monitoring. Future research can expand this work by incorporating larger datasets, hybrid optimization techniques, and transfer learning approaches to further improve model generalization and scalability.